\documentclass{article}
\usepackage{spconf,amsmath,amssymb,amsfonts,graphicx,hyperref}
\usepackage{cite}
\usepackage{algorithm}
\usepackage{algorithmic}
\usepackage{booktabs}
\usepackage{multirow}
\usepackage{array}
\usepackage{float}     
\usepackage[section]{placeins}  
\title{A Path Integral Surrogate for Multi-Step Gradient Inversion in Federated Learning}

\name{Agnivo Ghosh \qquad Saumik Bhattacharya}
\address{Department of Electronics and Electrical Communication Engineering\\
Indian Institute of Technology Kharagpur, India\\
\texttt{agnivo200325@kgpian.iitkgp.ac.in, saumik@ece.iitkgp.ac.in}}

\begin{document}
\ninept
\maketitle

\begin{abstract}
Federated learning lets many clients train a shared model together without ever sending their private data to a central server. Each client shares only a model update, and this update should reveal far less about the client than its raw training examples would. This premise is what protects the privacy of the clients. Gradient inversion attacks challenge it directly by trying to reconstruct a client's private input images from the single update it shared. Under FedAvg, a client's update accumulates several local training steps, so the server sees only the two endpoints of a hidden weight trajectory. Recent gradient inversion attacks fit a surrogate model along the path between these two endpoints but they still read its gradient at a single point. We propose the Path-Integral Surrogate Model Extension (PI-SME) which treats the accumulated update as a path integral of the gradient field and approximates it by Gauss--Legendre quadrature over several nodes along a learnable B\'ezier path. On CIFAR-100 and FEMNIST images across a range of trajectory lengths and class-restricted batches PI-SME reconstructs the private inputs more faithfully than the strongest surrogate baseline on several inversion metrics and the matching loss. 
\end{abstract}


\section{Introduction}\label{sec:intro}
Federated learning trains a shared model across many clients without collecting their data on a central server~\cite{fedavg,kairouz}. A server holds a global model and repeatedly sends it to participating clients. Each client owns a private dataset. Given a batch $(X,Y)$ of input images $X$ and labels $Y$, the client trains a model $f_{\mathbf{w}}$ with weights $\mathbf{w}$ under a loss $\ell$. It then returns only the resulting model update. The server aggregates these updates to form a new global model and repeats the process over multiple communication rounds. Since raw samples remain on the clients, federated learning is used in settings where privacy or data-governance requirements discourage central data collection, including mobile keyboard prediction~\cite{hard}, clinical modeling across hospitals~\cite{rieke}, fraud detection across financial institutions~\cite{yang}, and intelligence on edge and IoT devices~\cite{nguyen}.

The predominantly used federated learning algorithm is FedAvg~\cite{fedavg}. In FedAvg, each client performs several local SGD steps before returning its updated model to the server. The server then aggregates the client models using a weighted average. FedSGD is the one-step special case in which each client returns the gradient at the current global model rather than an update accumulated over multiple local steps. Sharing updates instead of data is intended to protect client privacy. This protection is further supported by defenses such as secure aggregation~\cite{bonawitz} and differential privacy~\cite{dpsgd}. Gradient inversion attacks challenge this protection by reconstructing private inputs from a shared update. An honest-but-curious server optimizes a dummy batch $\hat{X}$ with the same shape as $X$ until its gradient matches the observed update. DLG~\cite{dlg} minimizes the Euclidean distance between the two gradients, while iDLG~\cite{idlg} first recovers the labels. IG~\cite{ig} uses a cosine matching objective together with a total-variation image prior $\mathrm{TV}(\hat{X})$, while GradInversion~\cite{gradinversion} scales to larger batches. Prior-enhanced attacks such as GI-NAS~\cite{ginas} and DGGI~\cite{dggi} incorporate architectural or diffusion image priors. These methods are primarily designed for single-step gradients. In contrast, FedAvg produces an update accumulated over multiple local steps, and the server observes only the endpoints: the initial global model and the model returned by the client after local training. Recent surrogate-model attacks such as SME~\cite{sme} and NL-SME~\cite{nlsme} fit a surrogate path between these endpoints but still evaluate its gradient at a single point.

\textbf{Our contribution} is to treat the accumulated update as a path integral of the gradient field along the client trajectory. We reconstruct this integral using numerical quadrature rather than a single gradient evaluation. We propose the Path-Integral Surrogate Model Extension (PI-SME), which evaluates the surrogate gradient at several nodes along a learnable B\'ezier path and combines these evaluations using Gauss--Legendre weights to approximate the accumulated update. To our knowledge, this is the first gradient inversion attack to model the multi-step update as a path integral and use multi-node quadrature for inversion. On CIFAR-100~\cite{cifar} and FEMNIST~\cite{leaf,emnist}, PI-SME improves over the strongest surrogate baseline across all reconstruction metrics. The largest gains occur on long trajectories and class-scarce batches, where the single-point approximation is weakest.

Section~\ref{sec:related} reviews the multi-step setting and surrogate baselines. Section~\ref{sec:method} presents PI-SME. Section~\ref{sec:results} reports the main results and benchmark comparisons. Section~\ref{sec:ablation} presents the ablation studies. Section~\ref{sec:conclusion} concludes the paper.
\section{Background and Related Work}\label{sec:related}
\subsection{Multi-step updates in FedAvg}

The initial gradient inversion attacks of Section~\ref{sec:intro} assume a single-step FedSGD gradient. In practice a client takes $T$ local SGD steps before reporting. The $T$ steps produce the weight sequence $\mathbf{w}_0,\mathbf{w}_1,\dots,\mathbf{w}_T$, of which the server observes only the endpoints $\mathbf{w}_0$ and $\mathbf{w}_T$ while the intermediate states $\mathbf{w}_1,\dots,\mathbf{w}_{T-1}$ stay hidden. With learning rate $\eta$, and writing each per-step gradient on the full batch for clarity as in~\cite{sme}, the observed accumulated update is
\begin{equation}\label{eq:accum}
\Delta\mathbf{w}=\mathbf{w}_0-\mathbf{w}_T=\eta\sum_{\tau=0}^{T-1}\nabla_{\mathbf{w}}\ell\!\left(f_{\mathbf{w}_\tau}(X),Y\right),
\end{equation}
a sum of the gradients evaluated at the $T$ states $\mathbf{w}_0,\dots,\mathbf{w}_{T-1}$ that precede each step. Its direction is the mean gradient along the trajectory,
\begin{equation}\label{eq:gbar}
\bar{\mathbf{g}}=\frac{\Delta\mathbf{w}}{\eta T}=\frac{1}{T}\sum_{\tau=0}^{T-1}\nabla_{\mathbf{w}}\ell\!\left(f_{\mathbf{w}_\tau}(X),Y\right).
\end{equation}
Since $\Delta\mathbf{w}$ sums gradients over $T$ distinct states, a single-step attack evaluated at $\mathbf{w}_0$ mismatches the update, and the mismatch grows with $T$.
\subsection{Surrogate-model attacks}

SME~\cite{sme} attacks the multi-step update by matching the gradient of a surrogate model to $\Delta\mathbf{w}$. It restricts the surrogate to the line between the observed endpoints, $\widehat{\mathbf{w}}=\alpha\mathbf{w}_0+(1-\alpha)\mathbf{w}_T$, and jointly optimizes $\alpha\in[0,1]$ and the dummy batch $\hat{X}$ so that the surrogate gradient $\mathbf{g}_{\mathrm{sur}}=\nabla_{\mathbf{w}}\ell(f_{\widehat{\mathbf{w}}}(\hat{X}),Y)$ aligns with $\Delta\mathbf{w}$ under the cosine matching loss of~\cite{ig},
\begin{equation}\label{eq:lsim}
\mathcal{L}_{\mathrm{sim}}(\mathbf{g}_{\mathrm{sur}},\Delta\mathbf{w})
=1-\frac{\langle\mathbf{g}_{\mathrm{sur}},\Delta\mathbf{w}\rangle}
{\|\mathbf{g}_{\mathrm{sur}}\|\,\|\Delta\mathbf{w}\|},
\end{equation}
where $\Delta\mathbf{w}$ is the observed update of Eq.~\eqref{eq:accum}, $\langle\cdot,\cdot\rangle$ the inner product, and $\|\cdot\|$ the Euclidean norm. SME minimizes $\mathcal{L}_{\mathrm{sim}}$ over $\alpha$ and $\hat{X}$. NL-SME~\cite{nlsme} replaces the straight-line surrogate with a quadratic B\'ezier curve
\begin{equation}\label{eq:bezier}
\mathbf{w}(t)=(1-t)^2\mathbf{w}_0+2(1-t)t\,P_1+t^2\mathbf{w}_T,\qquad t\in[0,1],
\end{equation}
with a learnable control point $P_1$ and a learnable position $t$, and matches the gradient $\nabla_{\mathbf{w}}\ell(f_{\mathbf{w}(t)}(\hat{X}),Y)$ evaluated at that single position. NL-SME further introduces a learnable calibration vector $d$ that rescales the surrogate gradient coordinatewise before matching, which lowers $\mathcal{L}_{\mathrm{sim}}$ by up to an order of magnitude over SME~\cite{nlsme}. SME and NL-SME both read the surrogate gradient at a single state of the path.
\section{Proposed Method: PI-SME}\label{sec:method}
The update $\Delta\mathbf{w}$ accumulates gradients along the client's local trajectory, so up to the step size it is a discrete line integral of the gradient field. The quantity to reconstruct is therefore this path integral, not any single gradient. SME and NL-SME approximate the integral by reading the gradient at a single surrogate state, which replaces the whole path by one sample and leaves an error that grows as the trajectory lengthens with $T$. A single sample is also insufficient in principle. Matching $\bar{\mathbf{g}}$ from one state requires that state's gradient to be parallel to $\bar{\mathbf{g}}$, the average of the per-step gradients. For a scalar function the mean value theorem guarantees such a point, but it does not carry over to the vector-valued, high-dimensional gradients attacked here~\cite{rudin}. The path-consistency test in Section~\ref{subsec:mvt} confirms this on real trajectories, where the best single point is measurably worse than a quadrature over several nodes.

We therefore treat the reconstruction of $\Delta\mathbf{w}$ as a quadrature problem and approximate the path integral with $m$ fixed nodes along the path. Increasing the node count $m$ pulls the surrogate direction toward the true accumulated update and recovers more of it the longer and harder the trajectory is to match. Our method, the Path-Integral Surrogate Model Extension (PI-SME), realises this. Each node costs one backward pass, the dominant cost of the attack. The $m$ nodes are selected using Gauss--Legendre quadrature, which provides a high-order approximation with $m$ gradient evaluations and is exact for polynomial integrands of degree up to $2m-1$.

\subsection{Path-integral surrogate}
Modelling the unknown trajectory by the B\'ezier curve $\mathbf{w}(t)$ of Eq.~\eqref{eq:bezier}, the surrogate counterpart of $\bar{\mathbf{g}}$ is the path average
\begin{equation}\label{eq:integral}
I(\hat{X},P_1)=\int_0^1\nabla_{\mathbf{w}}\ell\!\left(f_{\mathbf{w}(t)}(\hat{X}),Y\right)dt.
\end{equation}
We evaluate this integral with an $m$-node Gauss--Legendre rule~\cite{gaussquad}, which for a continuous scalar integrand $h$ approximates
\begin{equation}\label{eq:gl}
\int_{-1}^{1} h(\xi)\,d\xi \approx \sum_{i=1}^{m} w_i\,h(\xi_i),
\end{equation}
and is exact for polynomials of degree up to $2m-1$; the nodes $\xi_i\in(-1,1)$ are the roots of the degree-$m$ Legendre polynomial and the weights $w_i$ are the corresponding closed-form Gauss--Legendre weights~\cite{gaussquad}. We apply the rule componentwise to the vector-valued gradient integrand. Since the surrogate curve is parameterised on $t\in[0,1]$, the affine change of variables $\xi\mapsto\tfrac{1}{2}(\xi+1)$ maps each node and weight,
\begin{equation}\label{eq:glmap}
t_i=\tfrac{1}{2}\left(\xi_i+1\right),\qquad c_i=\tfrac{1}{2}\,w_i,
\end{equation}
where the factor $\tfrac{1}{2}$ is the Jacobian of the map. The mapped weights satisfy $\sum_{i=1}^m c_i=1$, giving nodes $\{t_i\}_{i=1}^m\subset[0,1]$, so that
\begin{equation}\label{eq:quad}
G(\hat{X},P_1)=\sum_{i=1}^{m}c_i\,\nabla_{\mathbf{w}}\ell\!\left(f_{\mathbf{w}(t_i)}(\hat{X}),Y\right)
\end{equation}
is a weighted average of the surrogate gradient at $m$ fixed points of the curve. The node count $m$ sets how finely the path is resolved, and each node adds one gradient evaluation per iteration.
\subsection{Calibration and objective}
PI-SME matches $G$ to $\Delta\mathbf{w}$ under the same calibration as NL-SME~\cite{nlsme}. Since the reliability of the observed update can vary across layers, per-coordinate rescaling allows less reliable components to be downweighted during matching. A learnable vector $d$ of the same dimension as $G$ scales the integrated gradient, $d\odot G(\hat{X},P_1)$, where $\odot$ is the elementwise product, and is pulled toward the all-ones vector $\mathbf{1}$ by $\mathcal{R}_d=\|d-\mathbf{1}\|_2^2$. Setting $d=\mathbf{1}$ recovers the honest matching loss since the calibration can lower the reported loss without improving the true alignment. The control point is kept near the straight line by $\mathcal{R}_p=\|P_1-\tfrac{1}{2}(\mathbf{w}_0+\mathbf{w}_T)\|_2^2$. The reconstruction $\hat{X}$ is regularized by the anisotropic total-variation prior
\begin{equation}\label{eq:tv}
\mathrm{TV}(\hat{X})=\sum_{i,j}\Big(\big|\hat{X}_{i+1,j}-\hat{X}_{i,j}\big|+\big|\hat{X}_{i,j+1}-\hat{X}_{i,j}\big|\Big),
\end{equation}
where $\hat{X}_{i,j}$ is the pixel at row $i$ and column $j$, summed over the batch and colour channels. The reconstruction objective is
\begin{equation}\label{eq:obj}
\begin{split}
\mathcal{L}(\hat{X},P_1,d)={}&\mathcal{L}_{\mathrm{sim}}\!\left(d\odot G(\hat{X},P_1),\,\Delta\mathbf{w}\right)\\
&+\lambda_{\mathrm{tv}}\mathrm{TV}(\hat{X})+\lambda_p\mathcal{R}_p+\lambda_d\mathcal{R}_d,
\end{split}
\end{equation}
where $\mathcal{L}_{\mathrm{sim}}$ is the cosine matching loss of Eq.~\eqref{eq:lsim}, $G$ the integrated gradient of Eq.~\eqref{eq:quad}, $\Delta\mathbf{w}$ the observed update of Eq.~\eqref{eq:accum}, and $\lambda_{\mathrm{tv}},\lambda_p,\lambda_d>0$ weight the prior and the two regularizers. We set $\lambda_{\mathrm{tv}}=10^{-2}$, $\lambda_p=10^{-3}$, and $\lambda_d=10^{-6}$, following the calibration of NL-SME~\cite{nlsme}. We optimize $\Theta=\{\hat{X},P_1,d\}$ jointly with Adam~\cite{adam} for $1000$ iterations, at learning rates $1$ for $\hat{X}$, $10^{-3}$ for $P_1$, and $2.5\times10^{-3}$ for $d$, starting from $\hat{X}\sim\mathcal{N}(0,I)$, $P_1=\tfrac{1}{2}(\mathbf{w}_0+\mathbf{w}_T)$, and $d=\mathbf{1}$, with $d$ clamped to $[0.1,10]$. Labels are recovered from the last-layer update and held fixed. Algorithm~\ref{alg:pisme} lists the procedure.
\begin{algorithm}[t]
\caption{PI-SME reconstruction.}
\label{alg:pisme}
\begin{algorithmic}[1]
\REQUIRE observed models $\mathbf{w}_0,\mathbf{w}_T$; labels $Y$; nodes $m$; iterations
$K$; weights $\lambda_{\mathrm{tv}},\lambda_p,\lambda_d$
\ENSURE reconstructed batch $\hat{X}$
\STATE Gauss--Legendre nodes and weights $\{(t_i,c_i)\}_{i=1}^m$ on $[0,1]$,\ $\sum_i c_i=1$;\ \ $\Delta\mathbf{w}\gets\mathbf{w}_0-\mathbf{w}_T$
\STATE init $\hat{X}\sim\mathcal{N}(0,I)$,\ \ $P_1\gets\tfrac{1}{2}(\mathbf{w}_0+\mathbf{w}_T)$,\ \ $d\gets\mathbf{1}$
\FOR{$k=0$ \TO $K-1$}
  \FOR{$i=1$ \TO $m$}
    \STATE $\mathbf{w}(t_i)\gets(1-t_i)^2\mathbf{w}_0+2(1-t_i)t_iP_1+t_i^2\mathbf{w}_T$
    \STATE $g_i\gets\nabla_{\mathbf{w}}\ell(f_{\mathbf{w}(t_i)}(\hat{X}),Y)$
  \ENDFOR
  \STATE $G\gets\sum_{i=1}^{m}c_i\,g_i$
  \STATE $\mathcal{L}\gets\mathcal{L}_{\mathrm{sim}}(d\odot G,\Delta\mathbf{w})+\lambda_{\mathrm{tv}}\mathrm{TV}(\hat{X})+\lambda_p\mathcal{R}_p+\lambda_d\mathcal{R}_d$
  \STATE $(\hat{X},P_1,d)\gets$ Adam update on $\nabla\mathcal{L}$
\ENDFOR
\RETURN $\hat{X}$
\end{algorithmic}
\end{algorithm}
\section{Results}\label{sec:results}
We evaluate on CIFAR-100~\cite{cifar} ($100$ classes, $32\times32$ RGB images) and FEMNIST~\cite{leaf,emnist} ($62$ classes, $28\times28$ grayscale images), chosen to probe inversion across different terrains, colour and grayscale images with search spaces of different size. The victim networks are small CNNs with two convolutional layers followed by two fully connected layers, as in~\cite{sme,nlsme}, and the node count is fixed at $m{=}7$ from the ablation in Section~\ref{sec:ablation}. We compare PI-SME against SME, NL-SME, and three further gradient inversion attacks, test whether extra optimization closes the NL-SME gap, and measure reconstruction under low-rank updates.
\subsection{Benchmark comparison}

\begin{figure}[t]
\centering
\includegraphics[width=0.28\columnwidth]{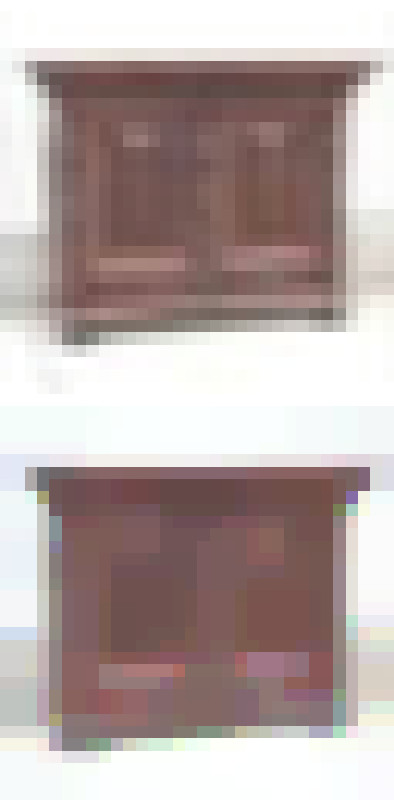}\hspace{2pt}%
\includegraphics[width=0.28\columnwidth]{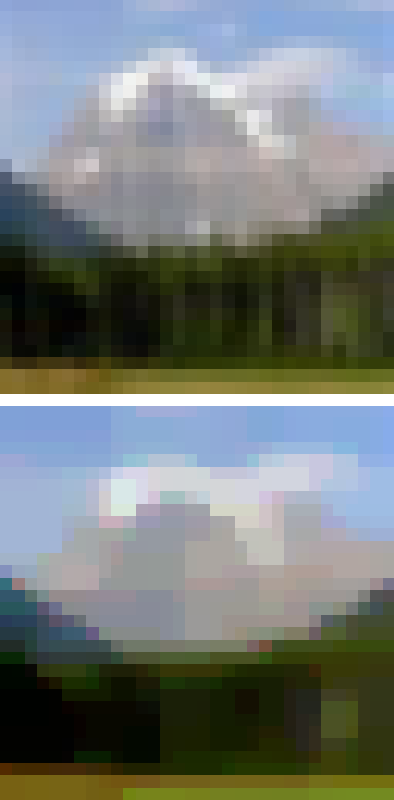}\hspace{2pt}%
\includegraphics[width=0.28\columnwidth]{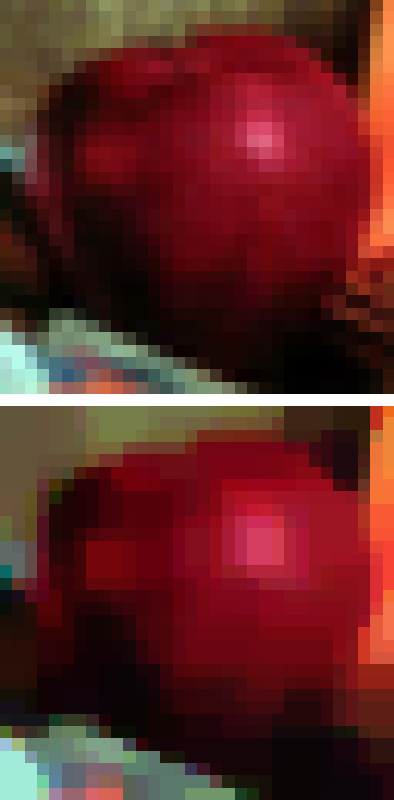}
\caption{Sample inversions using PI-SME on CIFAR-100; each column shows the original (top) and its reconstruction (bottom).}
\label{fig:samples}
\end{figure}

Figure~\ref{fig:samples} shows sample inversions using PI-SME on CIFAR-100. Table~\ref{tab:main} compares PI-SME with the surrogate baselines SME and NL-SME and with three further gradient inversion attacks, the single-step IG and the prior-enhanced GI-NAS and DGGI, on CIFAR-100 and FEMNIST. The single-step and prior-based methods trail the surrogate line by a wide margin on the multi-step update, since they are not built to follow an accumulated trajectory. Among the surrogate methods, PI-SME improves on NL-SME in every setting and on every metric. On CIFAR-100 the PSNR gain over NL-SME is about $0.8$~dB at $T{=}100$ and about $3$~dB at $T{=}250$; on FEMNIST it is about $1.2$~dB at both trajectory lengths, and SSIM~\cite{ssim}, FSIM~\cite{fsim}, LPIPS~\cite{lpips}, and $L_{\mathrm{sim}}$ move the same way.
\begin{table}[t]
\caption{Inversion metrics, mean $\pm$ variance for the PSNR (dB) column; best per column in bold. $E{=}20$ for $T{=}100$ and $E{=}50$ for $T{=}250$.}
\label{tab:main}
\centering
\footnotesize
\setlength{\tabcolsep}{3pt}
\renewcommand{\arraystretch}{0.92}
\begin{tabular}{c l c c c c c}
\toprule
& Method & PSNR (dB)$\uparrow$ & SSIM$\uparrow$ & FSIM$\uparrow$ & LPIPS$\downarrow$ & $L_{\mathrm{sim}}\downarrow$ \\
\midrule
\multirow{6}{*}{\rotatebox[origin=c]{90}{\shortstack{\textbf{CIFAR-100}\\\textbf{$T{=}100$}}}}
 & IG~(2020)      & $15.01\pm0.3156$ & $0.384$ & $0.719$ & $0.461$ & $0.1385$ \\
 & GI-NAS~(2025)  & $16.00\pm0.8537$ & $0.445$ & $0.732$ & $0.438$ & $0.1586$ \\
 & DGGI~(2025)    & $19.93\pm0.6229$ & $0.678$ & $0.825$ & $0.279$ & $0.0521$ \\
 & SME~(2023)     & $21.88\pm0.9726$ & $0.733$ & $0.861$ & $0.242$ & $0.0452$ \\
 & NL-SME~(2025)  & $25.77\pm0.2115$ & $0.855$ & $0.920$ & $0.163$ & $0.0032$ \\
 & PI-SME~(ours)  & $\mathbf{26.52\pm0.4288}$ & $\mathbf{0.868}$ & $\mathbf{0.926}$ & $\mathbf{0.149}$ & $\mathbf{0.0017}$ \\
\midrule
\multirow{6}{*}{\rotatebox[origin=c]{90}{\shortstack{\textbf{CIFAR-100}\\\textbf{$T{=}250$}}}}
 & IG~(2020)      & $12.28\pm0.2554$ & $0.221$ & $0.664$ & $0.517$ & $0.2681$ \\
 & GI-NAS~(2025)  & $12.70\pm0.3716$ & $0.246$ & $0.662$ & $0.509$ & $0.2914$ \\
 & DGGI~(2025)    & $18.60\pm0.4002$ & $0.576$ & $0.772$ & $0.380$ & $0.0592$ \\
 & SME~(2023)     & $18.26\pm0.3333$ & $0.563$ & $0.765$ & $0.389$ & $0.0625$ \\
 & NL-SME~(2025)  & $20.29\pm0.4908$ & $0.685$ & $0.827$ & $0.316$ & $0.0079$ \\
 & PI-SME~(ours)  & $\mathbf{23.20\pm0.3706}$ & $\mathbf{0.778}$ & $\mathbf{0.868}$ & $\mathbf{0.248}$ & $\mathbf{0.0036}$ \\
\midrule
\multirow{6}{*}{\rotatebox[origin=c]{90}{\shortstack{\textbf{FEMNIST}\\\textbf{$T{=}100$}}}}
 & IG~(2020)      & $17.09\pm0.3498$ & $0.524$ & $0.695$ & $0.365$ & $0.1617$ \\
 & GI-NAS~(2025)  & $17.31\pm0.2166$ & $0.562$ & $0.740$ & $0.300$ & $0.1929$ \\
 & DGGI~(2025)    & $21.33\pm0.4132$ & $0.771$ & $0.892$ & $0.114$ & $0.0408$ \\
 & SME~(2023)     & $22.27\pm0.4209$ & $0.790$ & $0.880$ & $0.173$ & $0.0338$ \\
 & NL-SME~(2025)  & $24.06\pm0.2576$ & $0.877$ & $0.956$ & $0.056$ & $0.0009$ \\
 & PI-SME~(ours)  & $\mathbf{25.34\pm0.3513}$ & $\mathbf{0.904}$ & $\mathbf{0.966}$ & $\mathbf{0.048}$ & $\mathbf{0.0006}$ \\
\midrule
\multirow{6}{*}{\rotatebox[origin=c]{90}{\shortstack{\textbf{FEMNIST}\\\textbf{$T{=}250$}}}}
 & IG~(2020)      & $13.64\pm0.3229$ & $0.397$ & $0.609$ & $0.429$ & $0.3102$ \\
 & GI-NAS~(2025)  & $13.62\pm0.4942$ & $0.435$ & $0.653$ & $0.369$ & $0.3332$ \\
 & DGGI~(2025)    & $20.50\pm0.2810$ & $0.739$ & $0.887$ & $0.118$ & $0.0484$ \\
 & SME~(2023)     & $21.35\pm0.3961$ & $0.754$ & $0.851$ & $0.213$ & $0.0482$ \\
 & NL-SME~(2025)  & $23.14\pm0.1863$ & $0.848$ & $0.945$ & $0.076$ & $0.0014$ \\
 & PI-SME~(ours)  & $\mathbf{24.37\pm0.2922}$ & $\mathbf{0.886}$ & $\mathbf{0.962}$ & $\mathbf{0.061}$ & $\mathbf{0.0006}$ \\
\bottomrule
\end{tabular}
\end{table}
\subsection{NL-SME under matched compute conditions}

A larger attack budget does not lift NL-SME inversion quality to PI-SME. On CIFAR-100 at $T{=}100$ its PSNR stays flat as the budget grows, at $25.76$, $25.63$, and $25.80$~dB for $1000$, $5000$, and $7000$ iterations. At $T{=}250$ it drops over the same three budgets, at $20.21$, $19.15$, and $19.26$~dB. Over this range its calibrated $L_{\mathrm{sim}}$ keeps falling while its honest matching loss (Eq.~\eqref{eq:obj} with $d{=}\mathbf{1}$) rises at $T{=}250$, from $0.0147$ to $0.0233$: the calibration vector $d$ absorbs the extra iterations instead of improving alignment with the observed update.
\subsection{Class deficiency conditions}

\begin{figure}[H]
\centering
\includegraphics[width=0.78\columnwidth]{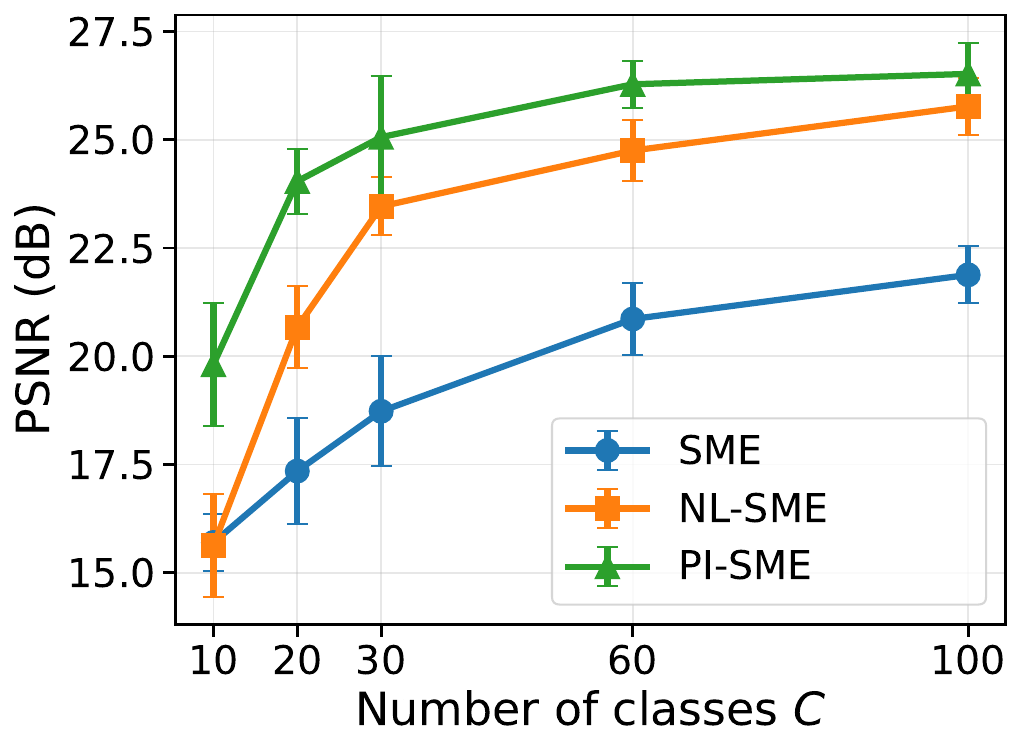}
\caption{PSNR versus the number of classes $C$ in the client batch on CIFAR-100, mean over five seeds.}
\label{fig:rank}
\end{figure}

For softmax cross-entropy the last-layer weight update $\Delta\mathbf{w}_{L}$ has rank
\begin{equation}\label{eq:rank}
\operatorname{rank}(\Delta\mathbf{w}_{L})\le\min(k,\,C-1),
\end{equation}
with $k$ the number of images in the batch and $C$ the number of classes, since the per-sample softmax residuals sum to zero across the $C$ logits. Fewer classes therefore leave a lower-rank signal to invert. Table~\ref{tab:main} shows that only SME and NL-SME come close to PI-SME, so we restrict the class-deficiency study to these three attacks. Figure~\ref{fig:rank} traces reconstruction as the CIFAR label set shrinks from $C{=}100$ to $C{=}10$. All three attacks lose quality as $C$ falls, and PI-SME degrades the slowest. Its PSNR margin over NL-SME widens from about $0.8$~dB at $C{=}100$ to about $4.2$~dB at $C{=}10$. At $C{=}10$ SME and NL-SME collapse close to each other while PI-SME stays several decibels ahead.
\section{Ablation Study}\label{sec:ablation}
\subsection{Node-count selection}
Table~\ref{tab:msweep} sweeps $m\in\{3,4,5,7,10,12\}$ on both datasets. The metrics change little across this range and a larger $m$ gives no monotone gain, with the best count varying by setting and seed. We fix $m{=}7$ for the remaining experiments, which sits near the best mean on both datasets while each added node costs one more gradient evaluation per iteration.
\begin{table}[t]
\caption{Node-count sweep for PI-SME: reconstruction PSNR (dB) across $m$
(mean over five seeds).}
\label{tab:msweep}
\centering\small
\setlength{\tabcolsep}{5pt}
\begin{tabular}{c c c c c}
\toprule
 & \multicolumn{2}{c}{CIFAR-100} & \multicolumn{2}{c}{FEMNIST} \\
\cmidrule(lr){2-3}\cmidrule(lr){4-5}
$m$ & $T{=}100$ & $T{=}250$ & $T{=}100$ & $T{=}250$ \\
\midrule
$3$  & $26.49$ & $22.97$ & $25.08$ & $24.00$ \\
$4$  & $26.63$ & $23.14$ & $25.21$ & $24.24$ \\
$5$  & $26.73$ & $23.05$ & $24.86$ & $24.34$ \\
$7$  & $26.52$ & $23.20$ & $25.34$ & $24.37$ \\
$10$ & $26.55$ & $23.10$ & $25.43$ & $24.34$ \\
$12$ & $26.72$ & $23.36$ & $25.09$ & $24.17$ \\
\bottomrule
\end{tabular}
\end{table}
\subsection{Path-consistency test}\label{subsec:mvt}
This experiment separates the numerical behaviour of the quadrature from the reconstruction problem. It tests whether integrating the surrogate gradient over several nodes of an optimized path aligns with the accumulated update more closely than the gradient read at any single state. The adversary is granted oracle access to the full sequence of true intermediate states $\{\mathbf{w}_\tau\}_{\tau=0}^{T}$ and to the private batch $X$, taken here from a CIFAR-100 client. No dummy data is optimized and the comparison is made purely between gradient directions and the true mean direction $\bar{\mathbf{g}}$ of Eq.~\eqref{eq:gbar}, scored by the honest cosine loss ($d{=}\mathbf{1}$).

The single-node reference measures the closest any one point on the true trajectory can come to $\bar{\mathbf{g}}$,
\begin{equation}\label{eq:dmvt}
\delta_{\mathrm{single}}=\min_{0\le\tau\le T}\mathcal{L}_{\mathrm{sim}}\!\left(\nabla_{\mathbf{w}}\ell(f_{\mathbf{w}_\tau}(X),Y),\;\bar{\mathbf{g}}\right),
\end{equation}
obtained by scanning all $T{+}1$ states, and is the strongest a single-state attack can reach even with full knowledge of the path. The quadrature counterpart evaluates the $m$-node estimate $G$ of Eq.~\eqref{eq:quad} on the B\'ezier path and reports the honest loss at the fitted control point,
\begin{equation}\label{eq:dquad}
\begin{split}
\delta_{\mathrm{quad}}(m)&=\mathcal{L}_{\mathrm{sim}}\!\left(G(X,P_1^\star),\bar{\mathbf{g}}\right),\\
P_1^\star&=\arg\min_{P_1}\ \mathcal{L}_{\mathrm{sim}}\!\left(G(X,P_1),\bar{\mathbf{g}}\right)+\lambda_p\mathcal{R}_p,
\end{split}
\end{equation}
where only the control point $P_1$ is fitted, from the two observed endpoints, and $\mathcal{R}_p$ is the control-point regularizer of Section~\ref{sec:method}. Table~\ref{tab:mvt} reports both quantities at two trajectory lengths. The quadrature loss lies below the single-node reference at every node count and at both lengths, even though $\delta_{\mathrm{single}}$ is computed over the entire trajectory while $\delta_{\mathrm{quad}}$ has access only to the two endpoints. Even a two-node quadrature already cuts the reference by roughly a factor of three, from $0.0242$ to $0.0079$ at $T{=}100$ and from $0.0446$ to $0.0136$ at $T{=}250$, and additional nodes lower it further before saturating near $m{=}3$. The advantage widens with the trajectory length, matching the growth of the reconstruction margin in Section~\ref{sec:results}. The best single state also moves along the path as $T$ grows, from $44\%$ of the way at $T{=}100$ to $71\%$ at $T{=}250$, so no fixed position is adequate across settings. 
\begin{table}[H]
\caption{Path-consistency test on a CIFAR-100 client: honest cosine loss against $\bar{\mathbf{g}}$.}
\label{tab:mvt}
\centering\small
\setlength{\tabcolsep}{5pt}
\begin{tabular}{l c c c c}
\toprule
 & $\delta_{\mathrm{single}}$ & $m{=}2$ & $m{=}3$ & $m{=}5$ \\
\midrule
$T{=}100$ & $0.0242$ & $0.0079$ & $0.0065$ & $0.0055$ \\
$T{=}250$ & $0.0446$ & $0.0136$ & $0.0113$ & $0.0119$ \\
\bottomrule
\end{tabular}
\end{table}
\section{Conclusion}\label{sec:conclusion}
PI-SME reframes multi-step gradient inversion as a quadrature problem. Because the FedAvg update is the accumulation of gradients along a client's local trajectory, the object worth matching is the path integral of the gradient field rather than the gradient at any one weight, and PI-SME approximates that integral with an $m$-node Gauss--Legendre rule along a learnable B\'ezier curve. 
The size of the gain tracks the difficulty of the setting. On the easy regimes PI-SME and the strongest surrogate baseline sit close together, but the margin over NL-SME widens as the client trains longer, from about $0.8$~dB at $T{=}100$ to about $3$~dB at $T{=}250$, and as the batch spans fewer classes, from about $0.8$~dB at $C{=}100$ to about $4.2$~dB at $C{=}10$. As the observed update drifts further from any single surrogate state, integrating along the trajectory recovers more of the signal. Choosing $m$ from a principled bound rather than a sweep and explaining the sharp fall all surrogate methods show on class deficient batches remain open as scope for future work.

\vfill\pagebreak

\bibliographystyle{IEEEbib}
\bibliography{bibliography}

\end{document}